\documentclass[letterpaper, 10pt]{article}
\usepackage{graphicx}
\usepackage{amsmath}
\usepackage{float}
\usepackage{amsfonts}
\usepackage[inner=0.75in, outer=0.75in, top=0.75in, bottom=1in, paperheight=11in, paperwidth=8.5in]{geometry}

\usepackage{epstopdf}
\usepackage{float}
\usepackage{amssymb}
\usepackage[utf8]{inputenc}
\usepackage[english]{babel}
\usepackage{amsthm}
\usepackage{color}
\usepackage{forest}

\usepackage{microtype}
\usepackage{url}
\usepackage{subfiles}
\usepackage{caption}
\usepackage{subcaption}
\usepackage{parskip}
\usepackage[section]{placeins}

\usepackage{tikz}
\usetikzlibrary{decorations.pathmorphing}
\usetikzlibrary{decorations.markings}
\usetikzlibrary{arrows.meta,bending}
\usetikzlibrary{arrows.meta}
\usepackage{bm}

\usepackage{xcolor}
\usepackage{framed}

\colorlet{shadecolor}{blue!20}

\usepackage[obeyFinal,textwidth=0.55in]{todonotes}
\usepackage[hidelinks]{hyperref}
\usepackage[capitalize]{cleveref}

\theoremstyle{definition}

\begin{document}	

\title{\bf On generating parametrised structural data using conditional generative adversarial networks}	
\author{G.\ Tsialiamanis, D.J.\ Wagg, N.\ Dervilis \& K.\ Worden\\
        Dynamics Research Group, Department of Mechanical Engineering, University of Sheffield \\
        Mappin Street, Sheffield S1 3JD, UK
	   }
	\date{}
    \maketitle
	\thispagestyle{empty}

\section*{Abstract}
A powerful approach, and one of the most common ones in structural health monitoring (SHM), is to use data-driven models to make predictions and inferences about structures and their condition. Such methods almost exclusively rely on the quality of the data. Within the SHM discipline, data do not always suffice to build models with satisfactory accuracy for given tasks. Even worse, data may be completely missing from one’s dataset, regarding the behaviour of a structure under different environmental conditions. In the current work, with a view to confronting such issues, the generation of artificial data using a variation of the generative adversarial network (GAN) algorithm, is used. The aforementioned variation is that of the \textit{conditional} GAN or cGAN. The algorithm is not only used to generate artificial data, but also to learn transformations of manifolds according to some known parameters. Assuming that the structure’s response is represented by points in a manifold, part of the space will be formed due to variations in external conditions affecting the structure. This idea proves efficient in SHM, as it is exploited to generate structural data for specific values of environmental coefficients. The scheme is applied here on a simulated structure which operates under different temperature and humidity conditions. The cGAN is trained on data for some discrete values of the temperature within some range, and is able to generate data for every temperature in this range with satisfactory accuracy. The novelty, compared to classic regression in similar problems, is that the cGAN allows unknown environmental parameters to affect the structure and can generate whole manifolds of data for every value of the known parameters, while the unknown ones vary within the generated manifolds.

\textbf{Key words: Structural health monitoring (SHM), machine learning, conditional generative adversarial networks (cGANs), environmental conditions, artificial data.}

\section{Introduction}
\label{sec:intro}

Structures are an irremovable part of modern society. Every part of modern life is connected to some structure. From buildings serving as houses and workplaces to vehicles facilitating human transport, structures prove to be essential elements of life. However, in order to deliver their function effectively, they should be monitored and their efficiency should be maintained. For this purpose, \textit{structural health monitoring} (SHM) has been developed. SHM is the discipline of inferring whether damage is present in a structure, what type of damage and where it is present, how severe it is and how much useful life remains for the structure \cite{rytter1993vibrational}. The most preferred way of performing such actions is through data-driven methods \cite{farrar2012structural}. Following such an approach, the results rely largely on the data. In some cases, trying to compensate for the lack of data, knowledge is induced into such \textit{black-box} models, forcing the model to learn in accordance with rules, that according to the user's knowledge are defined by the underlying physics problem. Combining knowledge and data in such a way leads to creation of \textit{grey-box} models.

A significant factor that affects structures, their behaviour and the potential appearance of damage is the environment within which they exist and interact. Most loads that a structure is excited by are because of the environment it operates in. Because of variations in their environmental conditions, the response of structures also varies. The behaviour of a structure may change in accordance with environmental factors such as temperature, humidity, direction of the load, magnitude of the load, etc. All these conditions change the way that a structure responds to excitation and sometimes such a change in the response may even be mistaken for existence of damage by algorithms that detect novelties in the behaviour of structures.

A quite simple data-driven approach to detecting existence of damage within a structure, is that of calculating novelty indices \cite{worden2003experimental}. The index is a distance metric between data acquired during the normal condition operation of the structure and a state of the structure that one wants to test for damage. Methods like this are efficient in some cases and able to indicate whether the structure behaves in a different way than normal. However, they are often vulnerable to false positive alerts regarding the existence of damage. The normal condition data are often represented only via their mean value and covariance matrix and the algorithm would be inefficient if extensive variations of normal condition structural behaviour existed.

Given a structure that operates within varying temperatures, its behaviour may change significantly as a function of the temperature. Lower temperatures can cause increases in stiffness of structural members and higher temperatures may have the opposite effect. Data can be acquired from a structure during a period in which its environment has a specific temperature. Following the novelty index approach would indicate damage existence if, during the period of the test state, a different temperature prevailed. But, even if a secondary and unknown environmental parameter variation appeared, such us change in humidity in specific parts of the structure, the novelty index approaches could also indicate existence of damage.

In the case of nonlinear structures, variance in normal condition features is also observed even as a result of variations in the characteristics of the loads. Linear structures would behave similarly under harmonic loads with different magnitudes; their \textit{frequency response function} (FRF) is not a function of the load but only of the structure's parameters (stiffness, damping, mass etc.). A structure with nonlinear elements, such as a \textit{Duffing oscillator}, has variations in behaviour, in terms of the FRF, according to the signal characteristics of the excitation force applied. Structures with much more complicated nonlinearities than a Duffing oscillator exist. Different kinds of nonlinearities (both geometric and in material behaviour) make the definition of normal condition characteristics a complicated task.

Approaches to deal with such issues have been developed. An example is presented in \cite{Worden1997}, where an auto-associative Network (AAN) \cite{kramer1991nonlinear} is used as a tool to learn the normal condition manifold of a structure. The network is trained using points that represent normal condition characteristics of the structure. The AAN learns from such points, how to project them in a lower dimensional space and then return them to the original space. Having been trained in such a way, a novelty index is defined as the reconstruction error of an input point. The error is defined as the Euclidean distance between the input and the output of the network. Points that lie on the manifold of normal condition operation, should have a small reconstruction error, since the AAN has learnt how to project onto this specific manifold. At the same time, points that diverge from this manifold would have a higher reconstruction error and this would indicate that damage may exist in the structure.

Following the aforementioned approach, a description of the normal condition manifold is achieved and anything that does not belong to that, is considered as a potential damage existence sign. The most trivial problem with the method, as with any fully data-driven method, is that the model, is only as good as the available data. Since prior physical knowledge is not induced by the user into the model, any state of the structure that is not represented somehow by the normal condition data, is considered as a potential damaged state. This fact may lead to false inference and start an unneeded maintenance procedure that would result in economic loss.

As an attempt to deal with such a problem, in the current work, an algorithm that learns transformations of manifolds is considered. The algorithm is that of the \textit{conditional generative adversarial network} (cGAN). The main difference of this algorithm, in contrast to the traditional \textit{(vanilla) GAN}, is that knowledge is introduced in the algorithm through a part of the input vector controlled by the user. By knowledge of environmental conditions, it is expected that, using the cGAN algorithm, a more concrete manifold of normal condition characteristics of the structure will be created. This opportunity may be exploited in order to build more robust SHM models but also in order to study and understand better the underlying physics of a problem in hand.

\section{Generative adversarial networks}
\label{sec:GANS}

\subsection{Introduction}
\label{sec:intro_GANs}

One of many frameworks for machine learning algorithms is that of training \textit{generative} models. Such models are able to generate samples that belong to the same probability density distribution as some given data and in the case of images this also translates into generating images that look real. A novel architecture for such models is the generative adversarial network \cite{goodfellow2014generative}. The novelty of the scheme is that of using two different neural networks in order to achieve the result of generating images that resemble reality. The two networks concerned are the \textit{generator} and the \textit{discriminator}. In the original work, the algorithm was focussed on images but throughout the current work, the algorithm shall be focussed on vector samples, since they are of more use in a dynamics or SHM project

Each of the two networks serves a different purpose throughout training, and their interaction yields the desired results. The discriminator on the one hand is optimised with a view to identifying whether a sample is real and comes from the original dataset or if it is fake/generated. On the other hand, the generator's training is focussed on ``fooling'' the discriminator by generating samples that the later will consider as real. Following such a scheme, throughout training, both networks perform their functionality better and the generator, eventually, is a network that will be generating real-like samples. The discriminator eventually performs in a similar way to the novelty detection algorithm presented in \cite{Worden1997}, learning the manifold of the data and indicating samples outside it as outliers.

As a general case, both networks can be multi-layer perceptron (MLP) networks. The generator $G$ takes as input a latent noise vector $\bm{z}$ from a probability distribution $p_{z}(\bm{z})$ and transforms it into a generated sample (or image) $G(\bm{z})$. The discriminator $D$ takes as inputs vectors (or images), and outputs the probability of them being real, i.e. $P_{\bm{x} \sim p_{data}} = D(\bm{x})$. The discriminator is optimised by setting as target values for the input samples their labels (``real'' or ``fake'') while the generator is optimised to make the discriminator classify the generated images as real, i.e. minimisation of $\log (1-D(G(\bm{z})))$. Therefore, the objective function of the described min-max game is given by,

\begin{equation} \label{eq:obj_fun}
    \min\limits_{G}\max\limits_{D}V(D,G)=\mathbb{E}_{\bm{x} \sim p_{data}(\bm{x})}[\log D(\bm{x})] + \mathbb{E}_{\bm{z} \sim p_{z}(\bm{z})}[\log(1 - D(G(\bm{z})))]
\end{equation}

\begin{figure}[h!]
    \centering
    \begin{tikzpicture}[scale=0.75, every node/.style={transform shape}]
        \definecolor{blue1}{RGB}{0, 128, 255}
        \node (1) at (0.0, 0.0) [draw, line width=0.5mm, fill=orange] {Noise, $\bm{z}$};
        
        \node (2) at (3.4, 0.0) [draw, line width=0.5mm, fill=blue1, minimum height=1.5cm, minimum width=2cm] {Generator};
        \draw[-{Latex[width=3mm, length=4mm]}, line width=0.5mm] (1) to (2);
        
        \node (3) at (6.8, 1.0) [draw, line width=0.5mm, fill=orange, minimum width=3.0cm] {\shortstack{Generated\\samples $G(\bm{z})$}};
        \node (4) at (6.8, -1.0) [draw, line width=0.5mm, fill=orange, minimum width=3.0cm] {\shortstack{Real\\ samples $\bm{x}$}};
        \draw[-{Latex[width=3mm, length=4mm]}, line width=0.5mm] (4.4, 0.0) to (5.3, 1.0);
        
        \node (5) at (11.2, 0.0) [draw, line width=0.5mm, fill=blue1, minimum height=1.5cm, minimum width=3.0cm] {Discriminator};
        
        \draw[-{Latex[width=3mm, length=4mm]}, line width=0.5mm] (8.3, 1.0) to (9.7, 0.0);
        \draw[-{Latex[width=3mm, length=4mm]}, line width=0.5mm] (8.3, -1.0) to (9.7, 0.0);
        
        \node (6) at (15.8, 0.0) [draw, line width=0.5mm, fill=orange, minimum height=1.5cm, minimum width=3.0cm] {Probability $D(G(\bm{z}))$};
        \draw[-{Latex[width=3mm, length=4mm]}, line width=0.5mm] (5) to (6);
        
    \end{tikzpicture}
    \caption{Vanilla GAN layout.}
    \label{fig:gan_layout}
\end{figure}
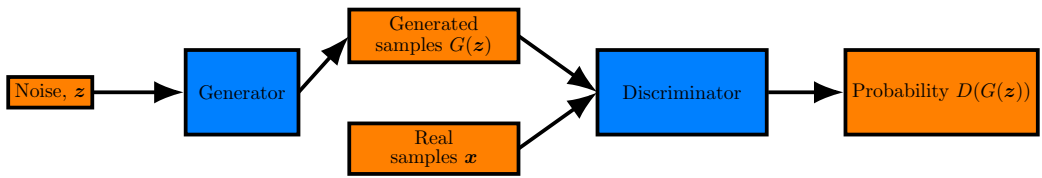

In practice, training is performed in two steps per epoch. During the first step, the discriminator is trained. Samples from the latent distribution $p(\bm{z})$ are sampled, and using them, the generator outputs samples. A second batch of samples is picked randomly from the original dataset and concatenated with the generated ones. Using the two batches, the discriminator is trained having as target labels, $1$ for the real samples and $0$ for the generated. During this phase, only the first term of the right hand side of equation (\ref{eq:obj_fun}) is used and its maximisation is sought. During the second phase, only samples from the latter distribution are sampled and this time the two networks are clipped together as shown in Figure \ref{fig:gan_layout}. This time, the trainable parameters of the discriminator are considered constants and they are not optimised. The target label set for training is $1$ for all samples, meaning ``real''. 

The error is back-propagated through the weights and biases of the generator. During this step, the second term of the right hand side of equation (\ref{eq:obj_fun}) is used and the objective is its minimisation. By the competitive procedure described, both networks get better at their respective objectives. In most cases, only the generator is utilised, with a view to generating samples, while the discriminator is considered an auxiliary network used only for training.

Properly trained GANs could prove quite useful for SHM purposes. In SHM acquiring enough data in order to train a machine learning model is sometimes hard and might be expensive. Therefore, using a GAN in order to augment a dataset with artificial data is a possibly valuable use of the algorithm. The GAN can also be used as a tool to implicitly enhance the performance of a neural network model, as described in \cite{mcdermott2018semi}, by exploiting unlabelled data. Most applications of GANs are found in image processing \cite{perarnau2016invertible, brock2016neural, zhang2019image}. Some of them have the quite impressive result of manipulating image features through the latter variables of the GAN. In some cases, GANs are used in order to fill gaps in images \cite{pathak2016context, li2017generative}. All these applications encourage the existing belief that generative models may be able to understand and recreate the physics of the underlying problems and should be exploited in order for humans to also gain further insight into problems in which they lack knowledge. Even an augmentation of data may assist in this task by providing more samples to study and understand underlying patterns and trends in the data.

\subsection{Conditional generative adversarial networks}
\label{sec:cGANs}
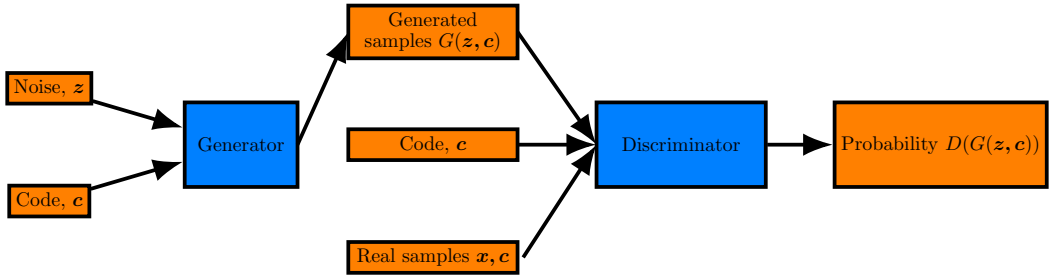
\begin{figure}[h!]
    \centering
    \begin{tikzpicture}[scale=0.75, every node/.style={transform shape}]
        \definecolor{blue1}{RGB}{0, 128, 255}
        \node (1) at (0.0, 1.0) [draw, line width=0.5mm, fill=orange] {Noise, $\bm{z}$};
        
        \node (A) at (0.0, -1.0) [draw, line width=0.5mm, fill=orange] {Code, $\bm{c}$};
        
        \node (2) at (3.4, 0.0) [draw, line width=0.5mm, fill=blue1, minimum height=1.5cm, minimum width=2cm] {Generator};
        \draw[-{Latex[width=3mm, length=4mm]}, line width=0.5mm] (1) to (2);
        
        \draw[-{Latex[width=3mm, length=4mm]}, line width=0.5mm] (A) to (2);
        
        \node (3) at (6.8, 2.0) [draw, line width=0.5mm, fill=orange, minimum width=3.0cm] {\shortstack{Generated\\samples $G(\bm{z, c})$}};
        \node (4) at (6.8, -2.0) [draw, line width=0.5mm, fill=orange, minimum width=3.0cm] {Real samples $\bm{x,c}$};
        
        \node () at (6.8, 0.0) [draw, line width=0.5mm, fill=orange, minimum width=3.0cm] {Code, $\bm{c}$};;
        
        \draw[-{Latex[width=3mm, length=4mm]}, line width=0.5mm] (4.4, 0.0) to (5.3, 2.0);
        
        \node (5) at (11.2, 0.0) [draw, line width=0.5mm, fill=blue1, minimum height=1.5cm, minimum width=3.0cm] {Discriminator};
        
        \draw[-{Latex[width=3mm, length=4mm]}, line width=0.5mm] (8.3, 2.0) to (9.7, 0.0);
        \draw[-{Latex[width=3mm, length=4mm]}, line width=0.5mm] (8.4, -2.0) to (9.7, 0.0);
        
        \draw[-{Latex[width=3mm, length=4mm]}, line width=0.5mm] (8.3, 0.0) to (9.7, 0.0);
        
        \node (6) at (15.8, 0.0) [draw, line width=0.5mm, fill=orange, minimum height=1.5cm, minimum width=3.0cm] {Probability $D(G(\bm{z, c}))$};
        \draw[-{Latex[width=3mm, length=4mm]}, line width=0.5mm] (5) to (6);
        
    \end{tikzpicture}
    \caption{Layout of a cGAN.}
    \label{fig:cgan_layout}
\end{figure}

Conditional generative adversarial networks are an attempt to control the output of the generator by conditioning on some variables. In contrast to the traditional GAN layout (Figure \ref{fig:gan_layout}), where the product of the generator is completely controlled by random noise $\textbf{z}$, the output of the generator is partially controlled by some vector $\textbf{c}$ (called a \textit{code} here-in, in parallel with the code introduced in the infoGAN \cite{chen2016infogan}). This code may be a continuous variable or a discrete/categorical one. Since the output of the generator depends on the code, so shall be the output of the discriminator. Following such a scheme, the layout of the cGAN is shown in Figure \ref{fig:cgan_layout}.

According to this approach, the discriminator learns that for each different value of $\textbf{c}$, a different decision boundary (or manifold) is defined. According to this realisation, the discriminator creates a different decision boundary in the feature space, as a function of $\textbf{c}$, for the outputs of the generator, forcing the latter to create samples that correspond to the values of the code (also given as inputs to the generator). In cases of discrete or categorical variables, the result is that the generator learns to generate samples belonging to different categories. In \cite{mirza2014conditional}, an illustration of this result is presented in the MNIST dataset; a dataset comprised of figures of hand-drawn digits between $0$ and $9$. By defining nine categorical variables and a one-hot encoding of the different classes, the generator is able to generate sample images in predefined classes, controlled by the code.

The use of continuous variables yields more interesting results. A continuous variable would force the decision boundary created by the discriminator, within the multidimensional feature space of the samples, to get gradually transformed according to the values of the code. The decision boundaries of the discriminator, in turn, force the generator to create samples within the region they define. Consequently the geometry of the generated manifolds is controlled by the code vector $\textbf{c}$. Following this procedure, what the generator has eventually learnt is the transformation of the manifold of the generated points, as a function of the code.

The algorithm may be exploited in order to generate artificial data as a function of some code vector. In the current work, it is exploited in order to learn the transformations of the aforementioned manifolds, as functions of the code. The most straightforward way that the algorithm might be used would be to fully control the class of the generated samples (as in the case of the MNIST dataset) and create artificial datasets according to one's needs. For SHM purposes, a more convenient use is that of learning how the manifold of data transforms according to some environemental parameter. 

As mentioned already, a common problem that arises in SHM is that the normal condition data are different for varying environmental conditions. Assuming that the normal condition data are not a single point but a manifold of points (as a result of noise or unknown environmental factors affecting the structure), their transformations as a function of known environmental parameters are desired. Learning such transformations would allow generating the normal condition manifold for unseen values of the code, understanding better the way that the structure behaves and creating a more complete dataset regarding its normal condition/undamaged state. In the following section, an application in a toy dataset is illustrated with a view to explaining the proposed algorithm and afterwards, an SHM application on simulated data is presented.

\section{Applications}
\label{sec:applications}

\subsection{Toy dataset}
\begin{figure}[h!]
    \centering
    \includegraphics[scale=0.7]{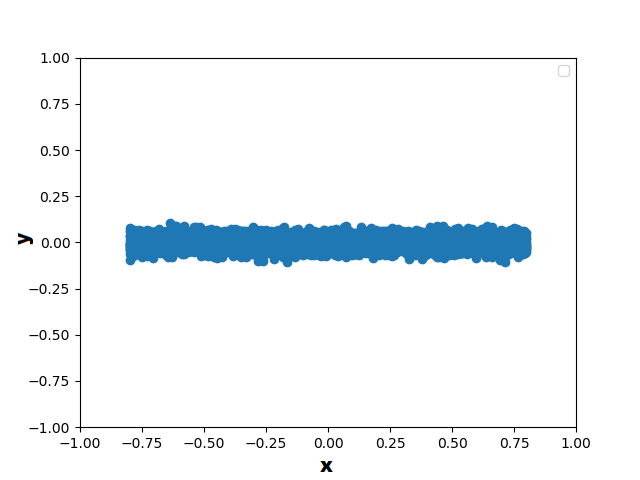}
    \caption{Base data, corresponding to angle of 0.}
    \label{fig:base_line}
\end{figure}

\begin{figure}[h!]
    \centering
    \includegraphics[scale=0.7]{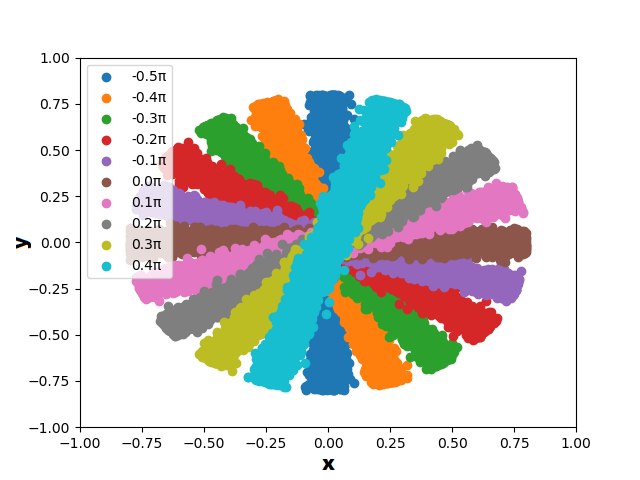}
    \caption{Collection of manifolds of the toy dataset.}
    \label{fig:all_toy_manifolds}
\end{figure}

Initially, in order to better illustrate the potential of the algorithm, a toy dataset was created. The data were two-dimensional and had $x$ values sampled uniformly in the range $[-1, 1]$ and $y$ values sampled from a normal distribution $y \sim \mathcal{N}(0, 0.03)$. The result is shown in Figure \ref{fig:base_line}. Afterwards, in order to create manifolds whose transformation was a function of a known parameter, the samples, shown in Figure \ref{fig:base_line}, were rotated using angles in the range of $[-\frac{\pi}{2}, \frac{\pi}{2}]$. The result is partially shown in Figure \ref{fig:all_toy_manifolds}.

\begin{figure}[h!]
    \centering
    \includegraphics[scale=0.7]{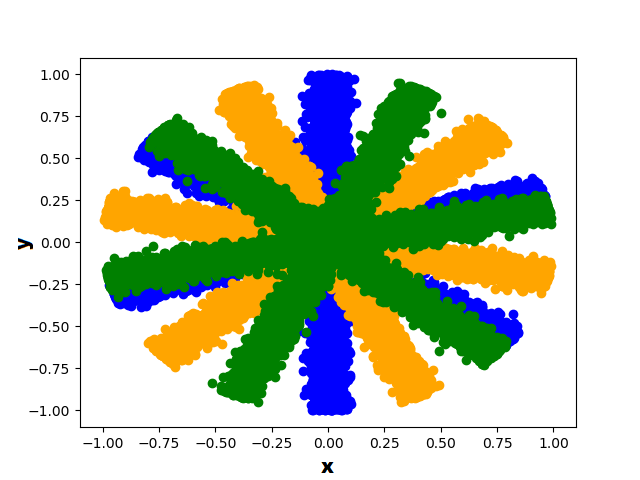}
    \caption{Points belonging to the training (blue), validation (orange) and testing (green) datasets.}
    \label{fig:train_test_val_toy_manifolds}
\end{figure}

The transformation of the initial manifold in this case is simply a rotation of the points around the point $(0, 0)$. In order to test if the cGAN would be able to learn such a transformation, a traditional scheme followed in machine learning was applied. Three datasets were created, a training, a validation and a testing one. The training dataset was created by generating points for $11$ values of the angle equally spaced in the interval $[-\frac{\pi}{2}, \frac{2\pi}{5}]$. The other two datasets were comprised of points for values of the angle also equally spaced in the same interval. Some of the manifolds within the created datasets are shown in Figure \ref{fig:train_test_val_toy_manifolds}. The training dataset should contain manifolds corresponding to the two outer values of the interval $[-\frac{\pi}{2}, \frac{2\pi}{5}]$ so that the transformation task is an interpolation across values of the angle parameter.

Since the loss function of a GAN does not directly represent the quality of the model \cite{goodfellow2014generative}, another training scheme has to be followed in order to have a metric of the performance of the algorithm on the validation set. The strategy used here is to calculate the \textit{Kullback–Leibler divergence} (KL divergence) \cite{kullback1997information} between the generated points for every validation code and the points in the validation dataset that correspond to each code. KL divergence is used to measure the distance between two probability distributions, so a probability distribution has to be defined for both the generated and the real points of each code. In order to define such distributions, kernel distributions \cite{silverman1981using} are fitted to both the real data and the generated. The divergence is then calculated between every pair of generated and real data for every code value. The model that has the lowest cumulative KL divergence in the validation set is considered the best one. The total KL divergence of a model is given by, 

\begin{equation} \label{eq:kl_divergence}
    \centering
    \sum_{n=1}^{n_{val}}D_{KL}(P_{i}||Q_{i}) = \sum_{n=1}^{n_{val}} \sum_{x \in X} P_{i}(x) \log(\frac{P_{i}(x)}{Q_{i}(x)})
\end{equation}
where the $P_{i}$ and $Q_{i}$ distributions are kernel density estimates fitted to the real and generated data, for the $i$-th code. The KL divergence is numerically computed on a grid in the feature space $X$ for discrete values of $x \in X$. The bandwidth of the kernel density function can be considered as a hyperparameter of the algorithm. In the current work, a reasonable value for the bandwidth was considered to be $0.2$, since all the data are normalised in the interval $[-1, 1]$ and a value equal to $\frac{1}{10}$ of the total range can sufficiently describe the distribution of data.

\begin{figure}[H]
    \centering
    \includegraphics[scale=0.45]{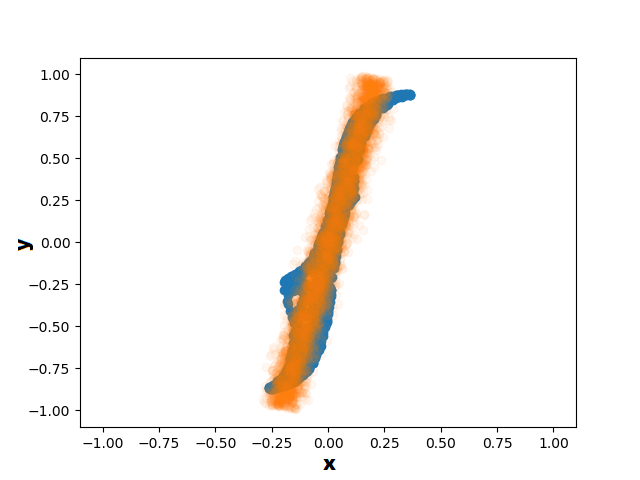}
    \caption{Real (orange) and generated (blue) points for code value in the testing dataset.}
    \label{fig:testing_codes_1}
\end{figure}

\begin{figure}[H]
    \centering
    \includegraphics[scale=0.45]{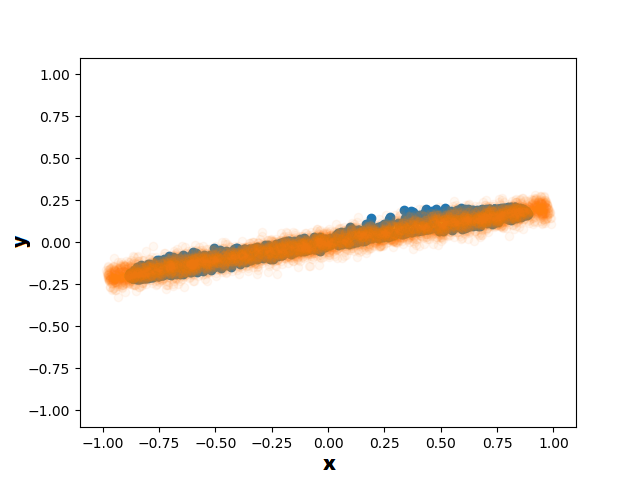}
    \caption{Real (orange) and generated (blue) points for code value in the testing dataset.}
    \label{fig:testing_codes_2}
\end{figure}

The generator was selected to be a three-layered neural network. The first layer was the input one with a total size of three (a two-dimensional noise $\textbf{z}$ plus an one-dimensional code $c$), a hidden layer, for which different sizes in the set $\{10, 20, ... 790, 800\}$ were tested, initiallising each randomly ten times. Eventually, a $500$-node sized one was found to yield the best results in terms of KL-divergence in the validation dataset. The output layer was a three-node layer. The activation function that performed best was \textit{hyperbolic tangent}. In every case the discriminator had the same size hidden layer as the generator. Following such a procedure, a model was defined whose performance was also tested on the testing dataset. The result for two code values belonging to the testing dataset are presented in Figures \ref{fig:testing_codes_1} \& \ref{fig:testing_codes_2}.

It is revealed that the algorithm has indeed learnt how the manifolds transform/rotate according to the value of the code. The overlap between the generated and the real points for code values that the cGAN has never ``seen'' before is quite satisfactory as shown in Figures \ref{fig:testing_codes_1} \& \ref{fig:testing_codes_2}. Another advantage of a cGAN approach is revealed in the specific toy example. The manifolds share regions within the two-dimensional space. There are regions common to every manifold, close to the center and manifolds with similar angle values share more points. If another variation of GAN were used, in order to achieve similar results, an extra axis would be needed in the data. This axis would be the variable that is known and that is induced as the code in the cGAN. Even doing so, the encoding of the new axis as a single latent variable of the GAN is not guaranteed.

\subsection{Generation of artificial parametrised data}

\begin{figure}[h!]
    \centering
    \begin{tikzpicture}
        \draw[line width=0.5mm] (-2,1) -- (-2,-1);

        \draw[line width=0.5mm] (-2, 1) -- (-2.2, 0.8);
        \draw[line width=0.5mm] (-2, 0.8) -- (-2.2, 0.6);
        \draw[line width=0.5mm] (-2, 0.6) -- (-2.2, 0.4);
        \draw[line width=0.5mm] (-2, 0.4) -- (-2.2, 0.2);
        \draw[line width=0.5mm] (-2, 0.2) -- (-2.2, 0.0);
        \draw[line width=0.5mm] (-2, 0.0) -- (-2.2, -.2);
        \draw[line width=0.5mm] (-2, -.2) -- (-2.2, -.4);
        \draw[line width=0.5mm] (-2, -.4) -- (-2.2, -.6);
        \draw[line width=0.5mm] (-2, -.6) -- (-2.2, -.8);
        \draw[line width=0.5mm] (-2, -.8) -- (-2.2, -1.0);
    
        \node (1) at (0.0, 0.0) [draw, line width=0.5mm, minimum width=1cm, minimum height=1cm] {$\bm{m_1}$};
        \node (2) at (2.5, 0) [draw, line width=0.5mm, minimum width=1cm, minimum height=1cm] {$\bm{m_2}$};
        \node (3) at (5.0, 0.0) [draw, line width=0.5mm, minimum width=1cm, minimum height=1cm] {$\bm{m_3}$};
        \node (4) at (7.5, 0.0) [draw, line width=0.5mm, minimum width=1cm, minimum height=1cm] {$\bm{m_4}$};
        \node (5) at (10.0, 0.0) [draw, line width=0.5mm, minimum width=1cm, minimum height=1cm] {$\bm{m_5}$};
        \node (6) at (12.5, 0.0) [draw, line width=0.5mm, minimum width=1cm, minimum height=1cm] {$\bm{m_6}$};
        
        \draw[thick, decoration={aspect=0.65, segment length=3mm,
             amplitude=0.2cm, coil}, decorate] (-2, 0) -- (-0.5, 0);
        
        \draw[thick, decoration={aspect=0.65, segment length=3mm,
             amplitude=0.2cm, coil}, decorate] (1) --(2);
        \draw[thick, decoration={aspect=0.65, segment length=3mm,
             amplitude=0.2cm, coil}, decorate] (2) --(3);
        \draw[thick, decoration={aspect=0.65, segment length=3mm,
             amplitude=0.2cm, coil}, decorate] (3) --(4);
        \draw[thick, decoration={aspect=0.65, segment length=3mm,
             amplitude=0.2cm, coil}, decorate] (4) --(5);
        \draw[thick, decoration={aspect=0.65, segment length=3mm,
             amplitude=0.2cm, coil}, decorate] (5) -- (6);

        \draw[-{Latex[width=3mm, length=4mm]}, line width=0.5mm] (0, 0.5) -- (0, 1.5) -- (1, 1.5);
        \node[] at (0.5, 2.0) {$F$};
        
        \node[] at (-1.25, 1.0) {$\bm{k_1}$};
        \node[] at (1.25, 1.0) {$\bm{k_2}$};
        \node[] at (3.75, 1.0) {$\bm{k_3}$};
        \node[] at (6.25, 1.0) {$\bm{k_4}$};
        \node[] at (8.75, 1.0) {$\bm{k_5}$};
        \node[] at (11.25, 1.0) {$\bm{k_6}$};
        
        \draw[line width=0.5mm] (0, -0.5) -- (0, -1.0) -- (-0.5, -1.0);
        \draw[line width=0.5mm] (-0.5, -0.9) -- (-0.5, -1.1);
        \draw[line width=0.5mm]  (-0.4, -1.2) -- (-0.6, -1.2) -- (-0.6, -0.8) --                             (-0.4, -0.8);
        \draw[line width=0.5mm] (-0.6, -1.0) -- (-0.9, -1.0) -- (-0.9, -1.6);
        \draw[line width=0.5mm] (-1.1, -1.6) -- (-.7, -1.6);
        
        \draw[line width=0.5mm] (2.5, -0.5) -- (2.5, -1.0) -- (2.0, -1.0);
        \draw[line width=0.5mm] (2.0, -0.9) -- (2.0, -1.1);
        \draw[line width=0.5mm]  (2.1, -1.2) -- (1.9, -1.2) -- (1.9, -0.8) --                             (2.1, -0.8);
        \draw[line width=0.5mm] (1.9, -1.0) -- (1.6, -1.0) -- (1.6, -1.6);
        \draw[line width=0.5mm] (1.4, -1.6) -- (1.8, -1.6);
        
        \draw[line width=0.5mm] (5, -0.5) -- (5, -1.0) -- (4.5, -1.0);
        \draw[line width=0.5mm] (4.5, -0.9) -- (4.5, -1.1);
        \draw[line width=0.5mm]  (4.6, -1.2) -- (4.4, -1.2) -- (4.4, -0.8) --                             (4.6, -0.8);
        \draw[line width=0.5mm] (4.4, -1.0) -- (4.1, -1.0) -- (4.1, -1.6);
        \draw[line width=0.5mm] (3.9, -1.6) -- (4.3, -1.6);
        
        \draw[line width=0.5mm] (7.5, -0.5) -- (7.5, -1.0) -- (7.0, -1.0);
        \draw[line width=0.5mm] (7.0, -0.9) -- (7.0, -1.1);
        \draw[line width=0.5mm]  (7.1, -1.2) -- (6.9, -1.2) -- (6.9, -0.8) --                             (7.1, -0.8);
        \draw[line width=0.5mm] (6.9, -1.0) -- (6.6, -1.0) -- (6.6, -1.6);
        \draw[line width=0.5mm] (6.4, -1.6) -- (6.8, -1.6);
        
        \draw[line width=0.5mm] (10, -0.5) -- (10, -1.0) -- (9.5, -1.0);
        \draw[line width=0.5mm] (9.5, -0.9) -- (9.5, -1.1);
        \draw[line width=0.5mm]  (9.6, -1.2) -- (9.4, -1.2) -- (9.4, -0.8) --                             (9.6, -0.8);
        \draw[line width=0.5mm] (9.4, -1.0) -- (9.1, -1.0) -- (9.1, -1.6);
        \draw[line width=0.5mm] (8.9, -1.6) -- (9.3, -1.6);
        
        \draw[line width=0.5mm] (12.5, -0.5) -- (12.5, -1.0) -- (12.0, -1.0);
        \draw[line width=0.5mm] (12.0, -0.9) -- (12.0, -1.1);
        \draw[line width=0.5mm]  (12.1, -1.2) -- (11.9, -1.2) -- (11.9, -0.8) --                             (12.1, -0.8);
        \draw[line width=0.5mm] (11.9, -1.0) -- (11.6, -1.0) -- (11.6, -1.6);
        \draw[line width=0.5mm] (11.4, -1.6) -- (11.8, -1.6);
        
        \node[] at (-1.25, -1.0) {$\bm{c_1}$};
        \node[] at (1.25, -1.0) {$\bm{c_2}$};
        \node[] at (3.75, -1.0) {$\bm{c_3}$};
        \node[] at (6.25, -1.0) {$\bm{c_4}$};
        \node[] at (8.75, -1.0) {$\bm{c_5}$};
        \node[] at (11.25, -1.0) {$\bm{c_6}$};

    \end{tikzpicture} 
    \caption{Mass-spring system.}
    \label{fig:mass_spring}
\end{figure}
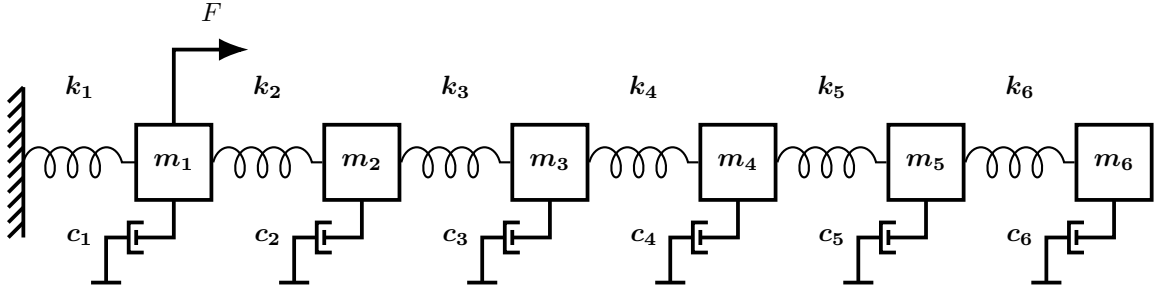

In order to test the potential of the algorithm in a different dataset, which is bound by some underlying physics, a simulated six-degree-of-freedom system was considered (Figure \ref{fig:mass_spring}). In the system, two different variables were chosen to affect its behaviour and to be causing variations in the characteristics. The first variable, the one known by the user and passed into the cGAN as prior knowledge, is the temperature. The temperature affects the stiffness of all springs. Higher temperatures causes stiffness reductions, while lower ones increase it. As a second varying environmental parameter, humidity was selected. Humidity in such a system could potentially affect the damping parameters of the system. Higher humidity could lead to higher damping coefficient $c_{i}$ and lower to the opposite. Temperature was selected to be varying in the interval $[20, 40]$ and humidity in the range $[80\%, 100\%]$.

\begin{figure}[h!]
    \centering
    \includegraphics[width=.55\textwidth]{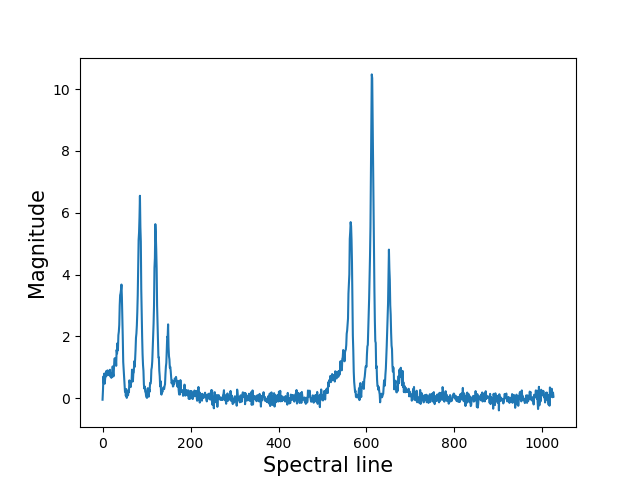}
    \caption{Example of simulated transmissibilities recorded.}
    \label{fig:typical_transes}
\end{figure}

The initial values of the structural parameters were $k_{i} = 10^{4}$, $c_{i} = 10$ and $m_{i} = 1$ ($i=1, 2...6$). In both cases, the relationship between the parameters and the environmental conditions was linear. Stiffness ($k_{i}$) increases or decreases linearly according to $k_{i} = k_{0} * (1 - \frac{T-30}{100})$, where $k_{0}$ is the initial value of the stiffness and $T$ is the temperature. Thereinafter, $1000$ simulations were performed for each one of $11$ discrete values of the temperature using a white noise excitation on mass $\#1$, as shown in Figure \ref{fig:mass_spring} and random values for humidity. Humidity was considered to affect the value of the damping parameters following the relationship $c = c_{0} * (1 - \frac{h-90}{100})$, where $h$ is the value of the humidity and $c_{0}$ the initial damping parameter values. For every simulation, the transmissibilities between the masses 1-2 and 2-3 were recorded, polluted with Gaussian noise (r.m.s. equal to $5\%$ of the signal's variance) and concatenated. A typical example of such a feature is shown in Figure \ref{fig:typical_transes}. 

\begin{figure}[H]
    \centering
    \includegraphics[scale=0.35]{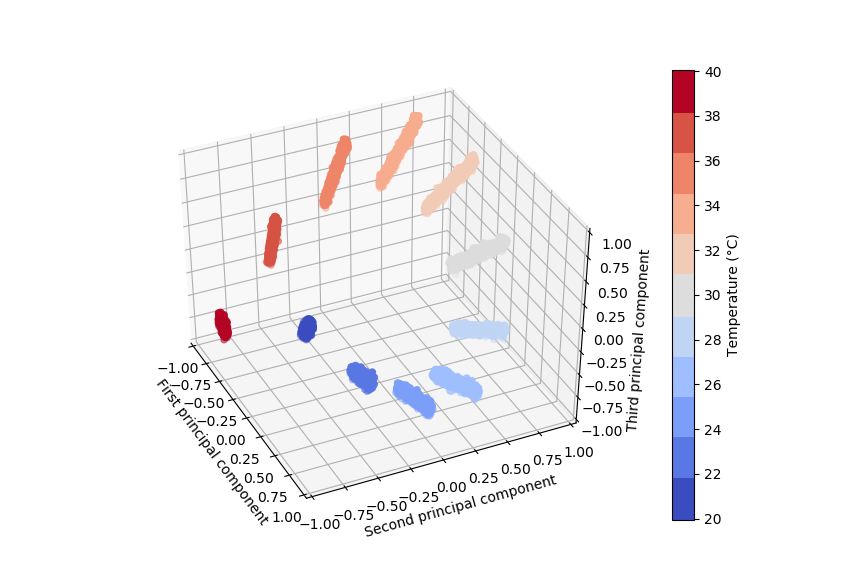}
    \caption{First three principal components of simulation data.}
    \label{fig:simulation_data}
\end{figure}

After collecting the transmissibilities, a \textit{principal component analysis} (PCA) \cite{wold1987principal} was performed, both in order to depict the data and to accelerate the training of the cGAN. The collection of generated points is shown in Figure \ref{fig:simulation_data}. The movement of the data within the feature space, according to the temperature, is revealed by the gradual colour change. The variation caused by the humidity is also visible, since each batch of points having the same temperature is spanned along a line segment. 

Among the 11 temperature values, one of them ($24^{\circ}$C), was considered to be the validation dataset and another one ($34^{\circ}$C) to give the testing. The highest and lowest values were included in the training dataset. Following the same procedure as before, the cGAN was trained and the criterion to choose which one performed better was once again the KL divergence between the real and generated points of the validation code/temperature. 

\begin{figure}[H]
    \centering
    \includegraphics[width=.50\textwidth]{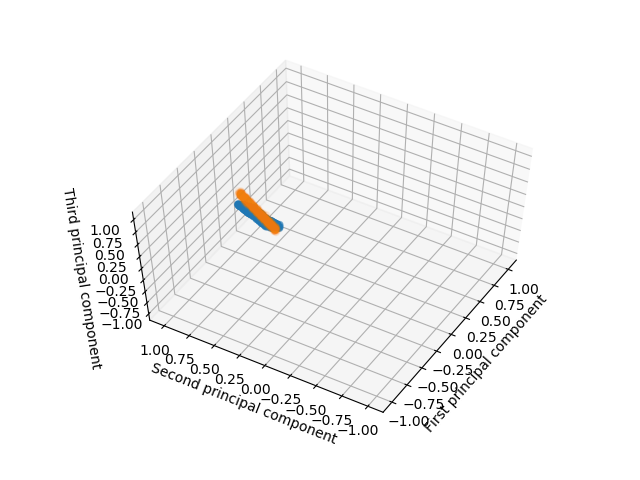}
    \caption{Real (orange) and generated (blue) points on the testing set for the temperature value of $34^{\circ}$C.}
    \label{fig:testing_temp_humidity}
\end{figure}

\begin{figure}[H]
    \centering
    \begin{subfigure}[b]{0.49\textwidth}
        \includegraphics[width=.90\textwidth]{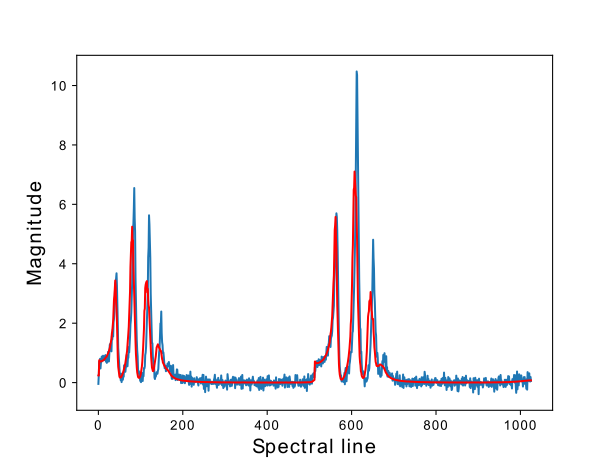}
    \end{subfigure}
    \begin{subfigure}[b]{0.49\textwidth}
        \includegraphics[width=.90\textwidth]{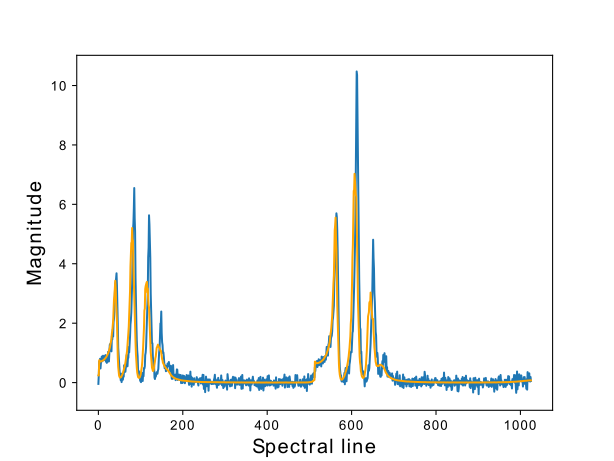}
    \end{subfigure}
    \caption{Transmissibility recorded from simulations with temperature $20^{\circ}$C (blue) compared to recorded transmissibility in $34^{\circ}$C (orange) and generated (red).}
    \label{fig:comparision_transes}
\end{figure}

\begin{figure}[h!]
    \centering
    \includegraphics[scale=0.40]{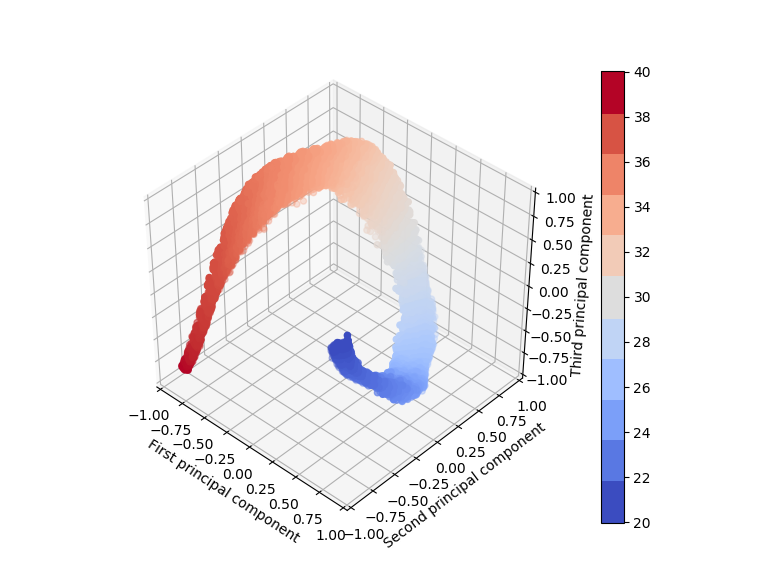}
    \caption{Manifold representing samples for all temperature values in the range $[20, 40]$.}
    \label{fig:manifold_all_temps}
\end{figure}

The performance on the testing set is shown in Figure \ref{fig:testing_temp_humidity}. The two batches (real and generated) are, as before, quite close. In order to illustrate also the results in terms of the transmissibilities, in Figure \ref{fig:comparision_transes} two tranmissibilities are shown, one generated using the testing code and one drawn from the testing dataset. For comparison, they are plotted along with a real transmissibility that corresponds to $20^{\circ}$C. It is clear that the cGAN has captured correctly the way that the diagram moves according to the temperature. A quite interesting application of such a model is that it may be used to generate the whole manifold of points for all intermediate temperatures that were not recorded for the structure. By generating points for many temperatures in the range $[20, 40]$ the resulting manifold is shown in Figure \ref{fig:manifold_all_temps}. Such a manifold provides a better description of the normal condition characteristics of the structure over many different temperatures. In the example presented here, the only unknown varying parameter was humidity but in any case there could be any number and the cGAN would be able to generate such manifolds for every value of the known code.

The cGAN is called upon to learn such movements and rotations (transformations) of the manifolds that arise because of the effect of temperature on the data. The convenience of the algorithm lies in the fact that a clear separation between known and unknown environmental parameters is performed. One needs to specify the values of the known parameters, and the algorithm is able to generate a range of plausible values (shown in Figure \ref{fig:simulation_data} as points with the same colour/temperature) of the feature vectors used for training. The aforementioned range is caused by the effect of unknown underlying factors, i.e. the variation in humidity herein. Having further knowledge about the factors, measuring more environmental parameters and using them as inputs in the algorithm, the aforementioned range, potentially, becomes smaller and the algorithm yields more confident predictions. Therefore, the code and noise vectors can be thought of as respective representatives of the existing knowledge about a problem and the \textit{epistemic uncertainty/lurking variables} affecting it.

\section{Conclusions}
\label{sec:conclusions}

The algorithm presented in the current work may be exploited in order to create data for unseen states of a structure. Data that correspond to discrete values of some parameter affecting the behaviour of the structure, are used to train the cGAN. Considering that the structure operates under known (or measured) and unknown (not measured or immeasurable) environmental parameters, the algorithm proposed can take both into account and generate artificial data for values of the known ones for which data are absent. This effect is achieved by learning transformations of manifolds according to the known environmental parameters. The method proposed in order to construct the whole manifold of potential states of a structure for varying environmental conditions provides a tool for understanding better the structural behaviour.

A major drawback of the described algorithm is the calculation of the KL divergence as an error metric in the validation dataset. The quantity of equation (\ref{eq:kl_divergence}) is calculated on a grid over the $\mathbb{R}^{n}$, where $n$ is the dimensionality of the data. Such a calculation is computationally inefficient. For example, having a grid with only $10$ points per axis would require computing the probability for every validation code in $10^{n}$ points. In the presented applications, the computation was performed in a three-dimensional space and so it was feasible, but for larger spaces it would not be. Nevertheless, since in SHM the monitored quantities can be low dimensional (or projected in lower-dimensional spaces, as was done here using PCA), the method still constitutes a viable tool.

The whole manifold of potential states may prove useful in situations where a novelty detection system is to be set up for monitoring. Apart from such applications, manifolds characterising structures may be even more useful in knowledge transfer situations. As proposed in \cite{PBSHM6}, a knowledge transfer scheme between structures of a population (even a heterogeneous one) can be applied in order to generate data for structures that one has not acquired. Under this framework, for each structure, the complete manifold of potential states, and not just parts of it, would make knowledge transfer much more applicable. In total, even the current method can be thought as a knowledge transfer method between states of the structure, parametrised by environmental conditions.

Finally, a useful and thought-invoking characteristic of the algorithm, is the way that knowledge and uncertainty are represented through the two parts of the input vector of the generator. Knowledge is represented through the code vector, which incorporates the known variables and parameters that affect the behaviour of the structure. Uncertainty, both epistemic and aleatory are encoded within the code vector. The cGAN generates data according to both, proving a convenient way of maybe propagating uncertainty in a fully data-driven framework.

\section{Acknowledgement}
\label{sec:ack}

The authors would like to acknowledge the support of the Engineering and Physical Science Research Council (EPSRC) and the European Union (EU). G.T. is supported by funding from the EU’s Horizon 2020 research and innovation programme under the Marie Skłodowska-Curie grant agreement DyVirt (764547). The other authors are supported by EPSRC grants EP/R006768/1, EP/R003645/1, EP/R004900/1 and EP/N010884/1.

\bibliographystyle{unsrt}
\bibliography{10213_tsi}

\end{document}